\documentclass[letterpaper]{article} 
\usepackage{aaai24}  
\usepackage{times}  
\usepackage{helvet}  
\usepackage{courier}  
\usepackage[hyphens]{url}  
\usepackage{graphicx} 
\usepackage{natbib}  
\usepackage{caption} 
\usepackage{algorithm}
\usepackage{algorithmic}
\usepackage{times}
\usepackage{epsfig}
\usepackage{graphicx}
\usepackage{amsmath}
\usepackage{amssymb}
\usepackage{multirow}
\usepackage{booktabs}
\usepackage{multirow}
\usepackage{siunitx}

\usepackage{hyperref}

\usepackage{xcolor}

\usepackage{newfloat}
\usepackage{listings}
\DeclareCaptionStyle{ruled}{labelfont=normalfont,labelsep=colon,strut=off} 
\floatstyle{ruled}
\newfloat{listing}{tb}{lst}{}
\floatname{listing}{Listing}
\title{A New Benchmark and Model for Challenging Image Manipulation Detection}

\author{
    Zhenfei Zhang\textsuperscript{\rm 1},
    Mingyang Li\textsuperscript{\rm 2},
    Ming-Ching Chang\textsuperscript{\rm 1}
}

\affiliations{
    \textsuperscript{\rm 1}Department of Computer Science, University at Albany, State University of New York, New York, USA, 12222\\
    \textsuperscript{\rm 2}Department of Bioengineering, McGill University, Montreal, QC, Canada, H3A 0E9\\
    zzhang45@albany.edu, mingyang.li@mail.mcgill.ca, mchang2@albany.edu 
}

\usepackage{bibentry}

\begin{document}

\maketitle

\begin{abstract}
The ability to detect manipulation in multimedia data is vital in digital forensics. Existing Image Manipulation Detection (IMD) methods are mainly based on detecting anomalous features arisen from image editing or double compression artifacts. All existing IMD techniques encounter challenges when it comes to detecting small tampered regions from a large image. Moreover, compression-based IMD approaches face difficulties in cases of double compression of identical quality factors. To investigate the State-of-The-Art (SoTA) IMD methods in those challenging conditions, we introduce a new Challenging Image Manipulation Detection (CIMD) benchmark dataset, which consists of two subsets, for evaluating editing-based and compression-based IMD methods, respectively. The dataset images were manually taken and tampered with high-quality annotations. In addition, we propose a new two-branch network model based on HRNet that can better detect both the image-editing and compression artifacts in those challenging conditions. Extensive experiments on the CIMD benchmark show that our model significantly outperforms SoTA IMD methods on CIMD. The dataset is available at: \url{https://github.com/ZhenfeiZ/CIMD}.
\end{abstract}

\section{Introduction}

With the advancement image editing and AI content generation, image editing, tampering and content synthesis are becoming common. However, the abuse of these technologies can bring in serious security and social impacts, including misinformation, disinformation, and deepfakes~\cite{Deepfake:TCS2021,Deepfake:Information_fusion2020}. {\bf Image Manipulation Detection (IMD)} methods that can accurately detect image manipulation regions are important in media forensics.

\begin{figure}[t]
\centerline{
  \includegraphics[width=\linewidth]{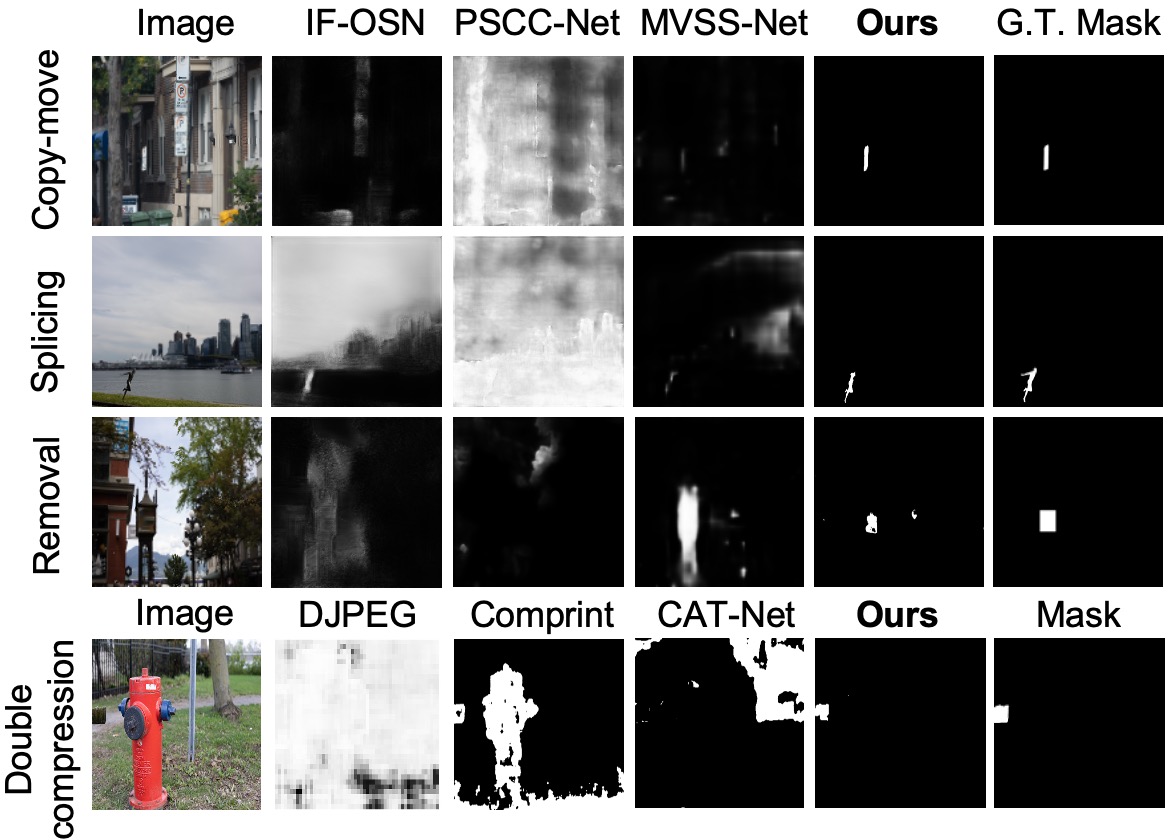}
}
\caption{Sample images of our dataset and comparison of image manipulation detection results with recent mainstream methods. The first three rows show manipulation of region copy-move, splicing and removal, respectively. The last row shows double-compressed splicing with the same Quality Factor (QF). Our method achieves the new state-of-the-art in detecting challenging manipulation cases.
}
\label{fig:front}
\end{figure}

\begin{figure*}[t]
\centerline{
  \includegraphics[width=\textwidth]{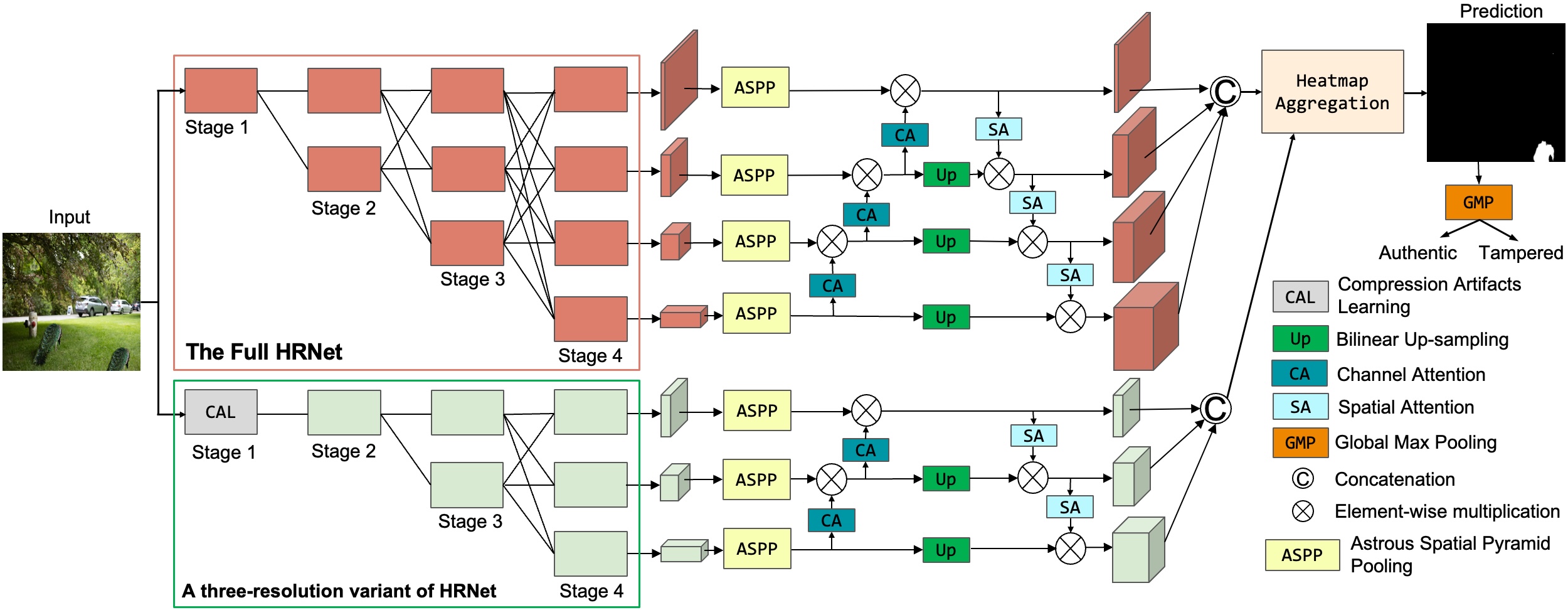}
}  
\caption{Overview of the proposed two-branch architecture. RGB stream can detect anomalous features, while frequency stream is able to learn compression artifacts by feeding the image to the compression artifacts learning model, as depicted in Fig.~\ref{fig:JPEG}. The ASPP in Fig.~\ref{fig:atten}(a) is appended to each of the outputs, and channel attention and spatial attention in Fig.~\ref{fig:atten}(b)(c) interactively perform between each scale output to improve the detection performance under small manipulation.
}
\label{fig:overall}
\end{figure*}

\begin{figure*}[t]
\centerline{
  \includegraphics[width=\linewidth]{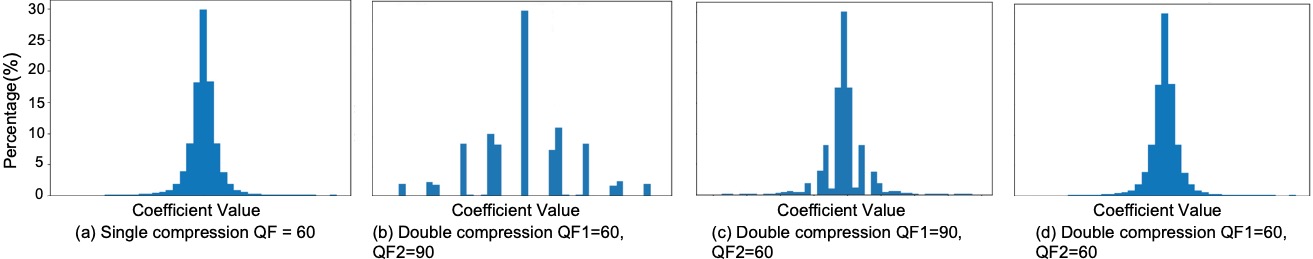}
}
\caption{DCT coefficient histograms from the (0,1) position generated from a raw image under different compression processes. The range of X-axis is [-20, 20].}
\label{fig:histogram}
\end{figure*}

There are three general types of image manipulation operations: (1) {\em region splicing}, where the content from one image is copied and pasted onto another image, (2) {\em region copy-move}, where an image regions is moved to another location within the same image, and (3) {\em region removal}, where parts of the image are erased and new contents are synthesized. To accurately detect these manipulations, some methods rely on detecting anomalous image region or texture features,
while others identify double compression artifacts.
While the State-of-the-Art (SoTA) IMD methods perform well on mainstream public IMD datasets,
they still face two challenges as shown in Fig.~\ref{fig:front}. First, existing IMD methods have general difficulties in detecting relatively small tampered regions, due to the data-driven design under limited visual information. Secondly, approaches detecting double compression inconsistencies with two different quantization matrices fall apart when the compression Quality Factor (QF) remains the same. This is because the use of identical Q-matrix can significantly suppress double compression artifacts. As shown in Fig.~\ref{fig:histogram}, methods in this category detect tampered regions by identifying missing histogram values arisen from the two compression processes. When the same QF is used, the histogram undergoes very small changes, making it hard to detect double compression. In summary, as the image tampering techniques improve increasingly fast, forensic problems are typically ill-defined, and IMD methods in general fall behind in research for challenging cases.


To address the issues and challenging conditions, we present a new two-branch IMD network incorporating both the RGB and frequency streams, such that both anomaly features and compression artifacts can be detected in a single framework. Our network adopts HRNet~\cite{HRNet:pami2020} as a feature extractor, with parallel processing at four different scales as in Fig.~\ref{fig:overall}. To more precisely pinpoint tiny tampering regions, we carefully designed the model by applying Atrous Spatial Pyramid Pooling (ASPP)~\cite{aspp:arXiv2017, dolg:iccv2021} and attention mechanism \cite{attention:nips2017, senet:cvpr2018}. For the frequency stream, we feed the backbone with quantized DCT coefficients, Q-matrix, and novel residual DCT coefficients from multiple recompressions to detect double compression artifacts. This design works regardless of different or identical QF's. To enhance the performance of the proposed two-branch model, we introduce an adaptive weighted heatmap aggregation design at the end, using soft selection to fuse the heatmaps generated by both branches. Our approach is distinct from the one used in \cite{higherhrnet:CVPR2020}, which relies on a simple averaging operation.

Datasets play a critical role in training and evaluating the performance of models. There is no publicly accessible datasets for challenging IMD cases. Existing datasets \cite{CASIA:2013, Coverage:icip2016, Culumbia:2009, Nist:wacv2019,MICC:tifs2011} exhibit a significant imbalance in the distribution of tampered images or contains only one image format, leading to an unreliable measurement of the overall detection capability of models. Additionally, some datasets~\cite{Defacto:eusipco2019, Imd:wacv2020} apply image tampering algorithms {\em e.g.}, \cite{inpainting:icip2014} to manipulate images in standard datasets such as MSCOCO, which raises concerns, as some IMD methods can rely on MSCOCO pre-trained backbones. 
In order to evaluate the effectiveness of IMD methods in challenging conditions, we propose a novel {\bf Challenging Image Manipulation Detection (CIMD)} dataset with new features. CIMD consists of two subsets for evaluations of image-editing-based and compression-based methods, respectively. 

The primary objective of the first subset is to evaluate the overall performance of image-editing-based methods in detecting small manipulation regions across all three types of manipulations. To ensure fair evaluation, we use raw images without any compression and ensure each type of manipulation contains the same number of samples. The main objective of the second subset is to assess the effectiveness of compression-based methods in detecting compression inconsistency using double-compressed images with identical QF. We created splicing manipulation images in which each double-compressed image was created using the same compression QF from 50-100. CIMD was taken and tampered with manually, ensuring high-quality image samples and annotations. 
We thus provide a reliable and accurate benchmark for evaluating the performance of image manipulation detection models. The availability of paired authentic and tampered images enables the comprehensive evaluation of a model's ability to identify manipulated images.
Contribution of this paper includes:

\begin{itemize}
\item We present a two-branch architecture incorporating RGB and frequency features for challenging image manipulation detection. To our knowledge, our model is the first approach to focus on detecting small tampered regions.

\item We introduced the pioneering compression artifacts learning model capable of detecting double-compression artifacts, regardless of whether the quantization factors (QFs) are different or identical. 

\item We introduce a new high-quality CIMD benchmark for evaluating the performance of SoTA IMD methods in challenging manipulations. 

\item Extensive experiments on CIMD demonstrate that the proposed approach outperforms the SoTA significantly in challenging image manipulation detection. 

\end{itemize}

\section{Related Work}

\subsection{Datasets for Image Manipulation Detection}

There are several datasets publicly available that are dedicated to image manipulation detection task. For example, the Columbia Dataset \cite{Culumbia:2009} contains uncompressed 363 splicing images of a low average resolution ($938 \times 720$). CASIA V1.0 and V2.0 \cite{CASIA:2013} were introduced for splicing and copy-move manipulation detection with no ground truth mask. Numerous datasets have been introduced only for copy-move tampering detection. For instance, the MICC \cite{MICC:tifs2011} features images mainly sourced from Columbia photographic image repository. Coverage \cite{Coverage:icip2016} is another copy-move only dataset includes 100 original-forged pairs with similar-but-genuine objects. The NIST~\cite{Nist:wacv2019} has presented benchmark manipulation datasets with multiple versions. Some large benchmark datasets, such as \cite{Defacto:eusipco2019} and \cite{Imd:wacv2020}, apply non-realistic questionable automatically forgeries methods \cite{inpainting:icip2014} to generate forgery images. In addition, to detect compression artifacts, \cite{Catnet:wacv2021} created five custom datasets that are double compressed using different unreported QFs. 

Most existing datasets in image manipulation detection only focus on a specific type of manipulation or exhibit a significant imbalance in the distribution of tampered types. This results in unreliable measurement of a model's overall detection capability. Furthermore, few datasets focus on challenging tampering detection. To address these limitations, we provide a novel dataset comprise two subsets: (1) Images with small manipulation regions, where each tampering type contains an equal number of instances, and (2) Images with spliced double-compression using identical QFs.

\begin{figure*}[t]
\centerline{
  \includegraphics[width=\textwidth]{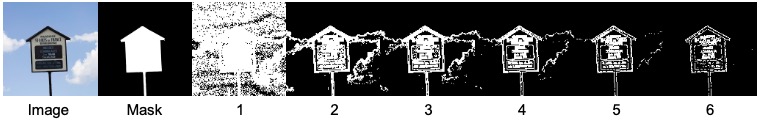}
}  
\caption{Visualization of DCT coefficients for each recompression for a repeatedly compressed image under QF 80. The number below shows recompression counts. Black pixels indicate unaltered DCT coefficients.
White pixels indicate the {\em unstable} region where DCT coefficients change after compression, which gradually focus on the tampered region as the count increases.}
\label{fig:recompression}
\end{figure*}

\subsection{Image Manipulation Detection}

Current methods for detecting image manipulation can be broadly classified into two categories that are distinguished by the manipulation artifacts they are designed to identify. Many technologies \cite{mvss:iccv2021, pscc:csvt2022, Rrunet:cvprw2019, ison:tifs2022, mantranet:cvpr2019,Span:eccv2020,Crcnn:icme2020,e2e:ieeea2020,objectFormer:cvpr2022} operate by detecting anomalous features. To accomplish this task, most of them utilize high-pass noise filters \cite{constrainedCnn:tifs2018, Highpass:iccv2019} to suppress content information. Other approaches \cite{Catnet:wacv2021,double_compression_detection:2018,comprint:2022} seek to identify compression inconsistencies in tampered images, as they assume that the compression QF's before and after manipulation differ. In addition to these two mainstream approaches, some researchers have directed their attention to camera-based artifacts, such as model fingerprints \cite{noiseprint:tifs2020,Splicebuster:wifs020,exif-sc:eccv2018,comprint:2022}.

In contrast to the methods mentioned above, our proposed approach employs a two-branch architecture that leverages both anomalous features and compression inconsistencies to detect image manipulation in more challenging conditions, which many current methods struggle to achieve. 
\section{The Challenging Image Manipulation Detection Dataset (CIMD)}
In this work, we aim to build a comprehensive validation dataset (CIMD) dedicated to small region forgery (less than 1.5\% on average) in both compressed and uncompressed scenarios. Our dataset are superior in image quality, image diversity, and forgery strategy. Two separate subsets have been introduced to evaluate image editing-based and compression-based methods, respectively.
\\\textbf{Collection.} We captured original images using Canon RP camera, encompassing both uncompressed TIFF and compressed JPG forgery-original image pairs. These captures were taken across highly diverse multi-season settings, characterized by intricate and sophisticated lighting conditions. Our intention was to offer an impartial and all-encompassing assessment of models within a real-life context.
\\\textbf{Two Disentangled Sub-Datasets.} We offer two subsets: the CIMD-Raw subset consists of pairs of original uncompressed TIFF images for the evaluation of image editing-based methods. The CIMD-Compressed subset encompasses splicing forgery and their corresponding original JPEG images with uniform quantization factors (QFs) ranging from 50 to 100. This subset evaluates the capability of compression-based models in detecting forgery under the same QF conditions.
\\\textbf{Processing and Tampering.} We used Photoshop 2023 (PS) to process and create tampering photos due to its popularity in other datasets mentioned in the related work section and its popularity in general public. 

\begin{figure*}[t]
\centerline{
  \includegraphics[width=\textwidth]{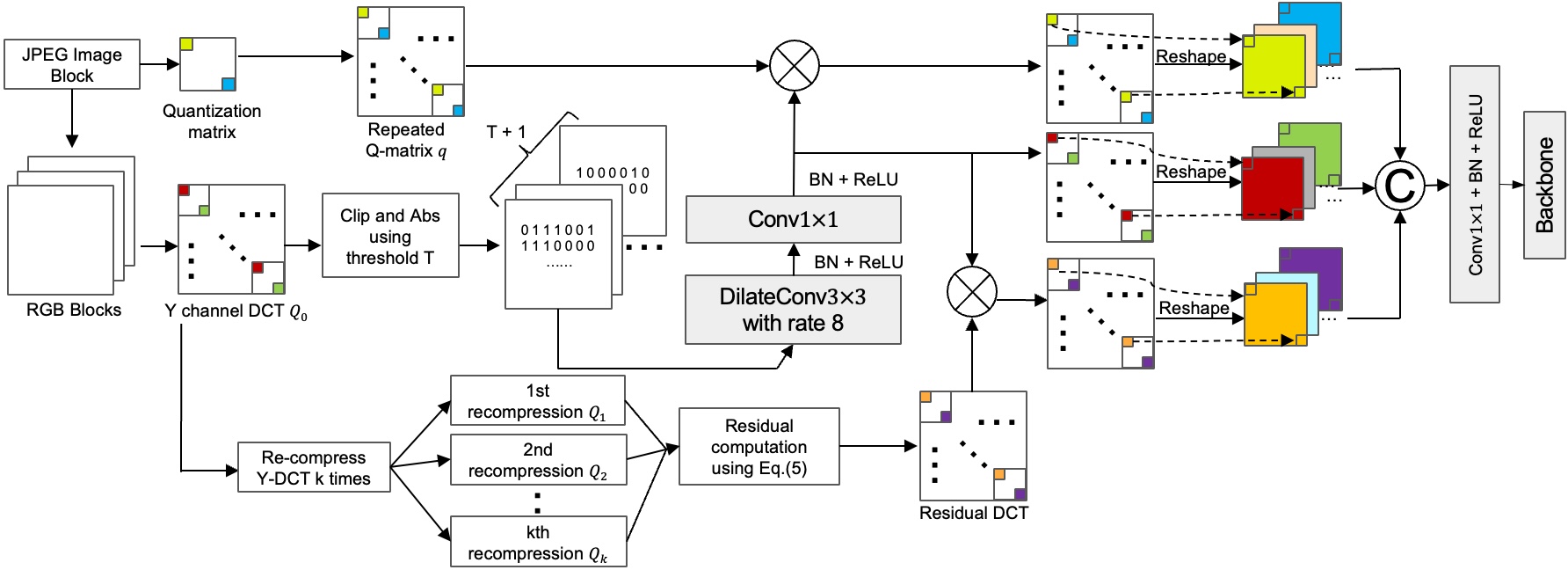}
}  
\caption{{The compression artifact learning module.} Three types ({\em de-quantized}, {\em quantized}, and {\em residual quantized}) of DCT features are fed into the backbone to learn double compression artifacts in cases whether the QFs are the same or not.}

\label{fig:JPEG}
\end{figure*}

\subsection{The CIMD-Raw (CIMD-R) Subset}

The CIMD-R benchmark provides a comprehensive evaluation of the image-editing-based models' performance in detecting small tampered copy-move, object-removal, and splicing forgeries on uncompressed images. The use of uncompressed images eliminates undesired compression artifacts on forgery region that can be otherwise sensed by neural networks, enabling a more true performance evaluation on out-of-detection. CIMD-R comprises 600 TIFF images, with a resolution of 2048 $\times$ 1365. Ground-truth masks are also provided. In addition, CIMD-R adopts a future-oriented approach by providing 16-bit image pairs that offer up to $2^{48}$ (trillions of) colors.
For copy-move manipulation, a part of an image is copied and pasted within the same image, followed by five post-processing methods: scaling, rotation, level/curve increasing, illumination changing, and color redistribution. In the case of removal manipulation, forged images are synthesized by removing the selected region from the image (via Content-Aware Fill in PS). Content-Aware Fill is widely used in several datasets~\cite{pspark2018double,Casiadong2013casia} and represents the PS's best guess to inpaint the object according to the surrounding region. For splicing forgery, regions from one image are copied and pasted into another source. Then, the same post-processing methods mentioned in copy-move are applied to make the forged region harmonious with its surroundings.


\subsection{The CIMD-Compressed (CIMD-C) Subset}

The CIMD-C benchmark is designed to evaluate the capability of compressed-based models in detecting double JEPG compression artifacts, where the primary and secondary compression has the same QFs. The dataset comprises 200 JPEG images with a resolution of 2048 $\times$ 1365, wherein the QF is uniformly distributed as $50 \leq QF < 100$.
Forgery images are generated akin to CIMD-R's splicing samples, with the distinction that the forged image is saved using the JPEG compression algorithm, employing the same QF as the original image. The original images were produced from RAW files ensuring that the original images are compressed for the first time, enhancing the dataset's credibility. In the forgery images, the background is double-compressed, while the tampered regions are single-compressed. Furthermore, the dataset also comprises binary masks and QF values utilized for compression, thereby augmenting its utility for further investigations into the effects of different QFs.

\section{The Proposed IMD Method}
\label{sec:method}

The two-branch architecture we propose enables the detection of both anomalous features and compression artifacts inspired by \cite{Catnet:wacv2021}. Furthermore, our model is effective for detecting small manipulation regions and identifying double compression traces that apply the same quantization matrix (Q-matrix). To achieve our research objectives, we adopted HR-Net \cite{HRNet:pami2020} as the backbone of our model, based on its ability to offer three-fold benefits. Firstly, the absence of pooling layers in HR-Net ensures that the features maintain high resolutions throughout the entire process. Secondly, the model processes features from different scales in parallel with effective information exchange, which is essential for capturing information of varying scales. Finally, the input size of HR-Net is ideally suited for DCT features. Since after processing by dilated convolution with a rate of 8, the size of the DCT feature is reduced to 1/8 of the input size, which is equivalent to the second stage resolution of HR-Net. 

\begin{figure*}[t]
\centerline{
  {\footnotesize (a)}
  \includegraphics[width=0.23\linewidth]{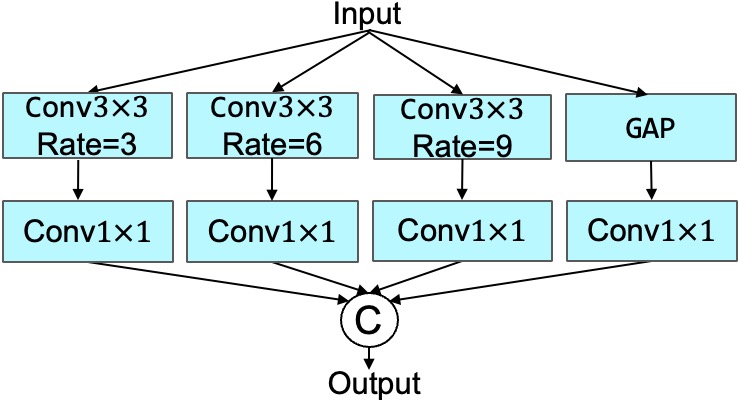}
  {\footnotesize (b)}
  \includegraphics[width=0.335\linewidth]{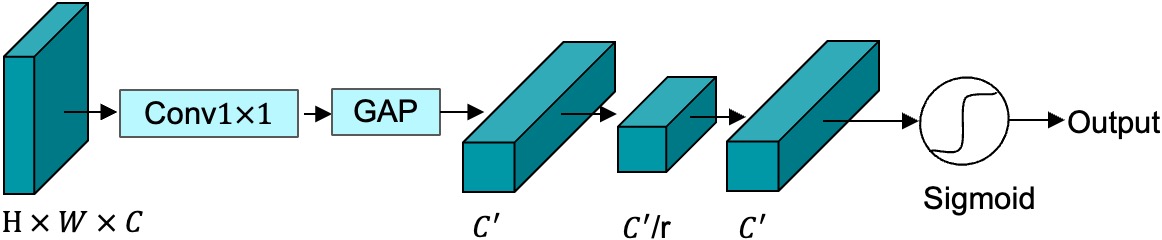}
  {\footnotesize (c)}
  \includegraphics[width=0.335\linewidth]{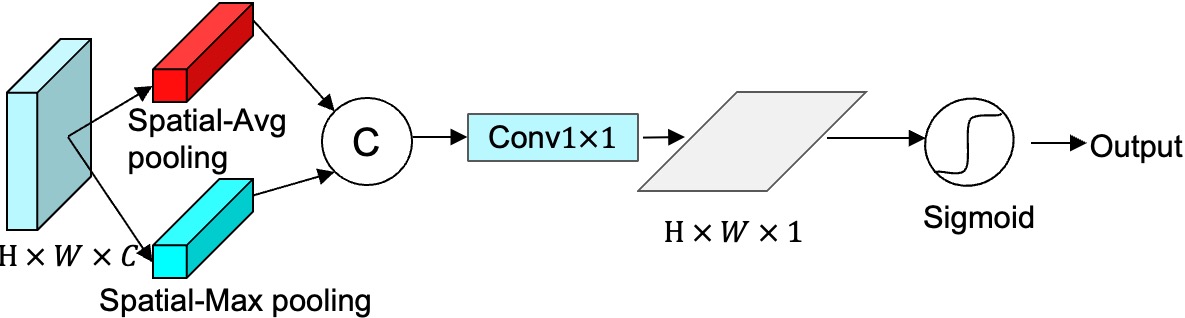}
}  
\caption{Detailed structure of the Atrous Spatial Pyramid Pooling (ASPP), channel attention and spatial attention. }

\label{fig:atten}
\end{figure*}

\subsection{Network Architecture}

The network architecture comprises two branches, one for detecting anomalous features and the other for identifying compression artifacts, as in Fig.~\ref{fig:overall}. For the RGB stream, the input image is fed to a full HR-Net, which learns the image editing traces from the visual content. In the frequency stream, the image is first input to the proposed compression artifact learning model shown in Fig.~\ref{fig:JPEG} to extract various DCT features. Subsequently, the DCT features are fed to a variant of the HR-Net, which operates at three different resolutions (1/8, 1/16, and 1/32).

To precisely pinpoint small tampering regions, we carefully designed our model using both Atrous Spatial Pyramid Pooling (ASPP) shown in Fig.~\ref{fig:atten}(a) and Attention Mechanism shown in Fig.~\ref{fig:atten}(b)(c). The ASPP captures long-range distance information via various receptive fields and handles scale variations. It consists of three dilated convolutional layers with different rates and a Global Average Pooling (GAP). The resulting features are concatenated and passed to a $1 \times 1$ convolution.

The starting point for designing an attention mechanism between each resolution output of HR-Net lies in the understanding that the four scale features extracted from HR-Net contain a diverse range of semantic and spatial information. Specifically, the high-resolution features contain more spatial content, whereas the low-resolution features carry more semantic responses. However, most prior methods simply do upsampling and concatenate these features for detection without adequately considering their inter-dependencies. The attention mechanism aims to fully leverage the information provided by each resolution and improve detection performance. Specifically, the approach utilizes channel attention from a bottom-up path and spatial attention from a top-down path, where two attention modules collaborate to enhance the features interactively. Through this approach, we seek to fully exploit the potential of each scale feature and improve detection performance.

We next describe how attention works interactively in the RGB stream, where the procedure is virtually identical to the frequency stream, with a different number of output resolution branches. Given a RGB input image $I$ with width $W$ and height $H$, $I \in \mathbb{R}^{H \times W \times 3}$, the HR-Net output features in four resolutions can be denoted as $F_{1} \in \mathbb{R}^{H/4 \times W/4 \times C_{1}}$, $F_{2} \in \mathbb{R}^{H/8 \times W/8 \times C_{2}}$, $F_{3} \in \mathbb{R}^{H/16 \times W/16 \times C_{3}}$ and $F_{4} \in \mathbb{R}^{H/32 \times W/32 \times C_{4}}$, and $C_{1}=48$, $C_{2}=96$, $C_{3}=192$, $C_{4}=384$ as default setting. The bottom-up channel attention feature are calculated using:
\begin{equation}
F_{n} = {\cal C}(F_{n+1})\odot F_{n}, \;\; n=1,2,3,
\label{eq:channel_atten}
\end{equation}
where ${\cal C}(\cdot)$ denotes the channel attention block in Fig.~\ref{fig:atten}(b) and $\odot$ represents element-wise multiplication. As $F_{4}$ contains the highest level of semantic information, it remains unchanged at the channel level.

For the detail of channel attention, the feature maps $F_{n+1}$ undergo an essential preliminary transformation through a $1 \times 1$ convolutional layer. This transformation is crucial to ensure that the number of channels between $F_{n+1}$ and $F_{n}$ is consistent, thereby enabling the element-wise multiplication to be performed effectively in the channel dimension. We set the transformed channel number as $C^{'}$. The transformed features are subsequently fed to a Global Average Pooling, denoted as $GAP(\cdot)$, followed by the excitation process $E(\cdot) = C^{'} \rightarrow C^{'}/r \rightarrow C^{'} $, $r = 4$). The channel attention is calculated as
$
{\cal C}(F) = \sigma \left( E(GAP(Conv_{1 \times 1}(F))) \right),
\label{eq:GAP}
$
where $\sigma(\cdot)$ is the Sigmoid activation function.

Following the application of bottom-up channel attention, the feature maps $F_{2}$, $F_{3}$, and $F_{4}$ are upsampled using the bilinear upsampling method to match the resolution of $F_{1}$. The spatial attention mechanism from the top-down pathway is then applied, which is given by:
\begin{equation}
F_{m} = S(F_{m - 1}) \otimes F_{m}, \;\; m = 2, 3, 4,
\label{eq:spatial_atten}
\end{equation}
where $S(\cdot)$ is the spatial-attention in Fig.~\ref{fig:atten}(c). As $F_{1}$ contains the richest spatial information, it remains unchanged at the spatial level. 
The spatial attention is calculated using the Spatial Max Pooling $P_{max}$ and Spatial Average Pooling $P_{avg}$ as
$
S(F) = \sigma \left( Conv_{1 \times 1}\left [ P_{max}(F);P_{avg}(F)\right ] \right),
\label{eq:spatial_operation}
$
where $\left [ ; \right ]$ denotes concatenation.

The feature maps of each branch, after undergoing upsampling and interactive attention, have the same resolution. These features are then concatenated together to form final features for adaptive weighted heatmap aggregation in inference stage. Our model generates two final heatmaps, which are aggregated through soft selection. Specifically, we employ bilinear feature upsampling to upscale the heatmap of the frequency stream to match the resolution of the RGB stream heatmap. Following this, we apply the Softmax activation function to the heatmaps, and then use Global Max Pooling (GMP), denoted as $GMP(\cdot)$, to select the main heatmap and its corresponding weight. This selection is based on higher values, which indicate a stronger localization response compared to the other heatmaps. We define the main and secondary heatmap using $h_{m}$ and $h_{s}$. Thus the weighted aggregated heatmap $h$ can be generated using: 
\begin{equation}
h = GMP(h_{m}) \cdot h_{m} + (1 - GMP(h_{m})) \cdot h_{s}.
\end{equation}
Finally, the same as \cite{mvss:iccv2021}, we apply a non-trainable GMP over the predicted binary mask to perform image-level detection, since image-level detection is highly related to pixel-wise prediction.

\subsection{JPEG Compression Artifacts Learning Model}

Our compression learning model aims to identify compression artifacts in double-compressed images, regardless of whether the primary and secondary compressions have the same QF or not. Several approaches attempt to detect inconsistencies in the DCT histogram, as illustrated in Fig.~\ref{fig:histogram}(b)(c). It should be noted that when double compression is performed using the same Q-matrix, histogram-based methods are not effective since there are very few compression inconsistencies, as shown in Fig.~\ref{fig:histogram}(d). Fortunately, some traces can still be detected even in such conditions. It was observed in \cite{double_detection_same_first:tifs2010} that when a JPEG image is repeatedly compressed using the same QF, the number of different quantized DCT coefficients between two consecutive compressions decreases monotonically. Several methods \cite{double_detection_same_cnn:icip2018,double_detection_same_error:tisf2014,double_detection_same_2022:csvt2022} leverage this evidence to determine whether an image has been single or double-compressed. In contrast to previous approaches, we investigate the feasibility of leveraging this trace to localize tampered regions in an image. 
Fig.~\ref{fig:recompression} shows that when a spliced image is created using the same QF, the manipulated region is singly compressed, however the background regions are doubly compressed. 
Consequently, when the image is repeatedly compressed, unstable quantized DCT coefficients gradually focus on the tampered area, while the authentic regions remain relatively stable. Based on this observation, we introduce a novel residual DCT map to guide the DCT features to better focus on the unstable regions for IMD.

Our method focuses only on Y-channel DCT map, as it is more sensitive to human eyes. Given a JPEG image, it is easy to obtain the Y-channel quantized DCT coefficients $Q_{0}$ and its corresponding Q-matrix from the JPEG file header. The Q-matrix is first repeated to have the same size as $Q_{0}$ and we set the repeated Q-matrix as $q$. Thus, We compute the $(k + 1)$th re-compression quantized JPEG coefficients $Q_{k+1}$ using the following equations sequentially:
\begin{equation}
\left\{
             \begin{array}{lr}
             D_{k} = Q_{k} \odot q \\
             B_{k} = IDCT(D_{k})\\
             I_{k+1} = RT(B_{k})\\
             Q_{k+1} = [DCT(I_{k+1}) \oslash q]
             \end{array},
\right.
\label{recompress}
\end{equation}
where $\oslash$ denotes element-wise division, $D$, $B$, $I$ and $Q$ represent de-quantized DCT coefficients, de-transformed blocks using inverse DCT, image blocks and quantized JPEG coefficients respectively. The subscripts of the variables in the above equations represent the number of recompressions and we experimentally set $k = 7$. $RT(\cdot)$ is rounding and truncation operation. $[\cdot]$ denotes to the rounding operation. Thus, the residual de-quantized DCT coefficients $R$ after k-times recompressions is defined as:
\begin{equation}
R = \frac{1}{k}\sum_{i = 1}^{k}(Q_{i} - Q_{i-1}).
\label{eq:residual_dct}
\end{equation}

\begin{table}[t]
\centerline{
\setlength{\tabcolsep}{1.4mm}
\footnotesize
  \begin{tabular}{lcccc}
    \toprule
      \multirow{2}{*}{Method}{} &
      \multicolumn{2}{c}{Pixel-level F1} &
      \multicolumn{2}{c}{Image Level} \\
      & Best & Fixed & AUC & Acc\\
      \midrule
    RRU-Net \cite{Rrunet:cvprw2019} &0.126 &0.103 & 0.500 & 0.500 \\
    CR-CNN \cite{Crcnn:icme2020} & 0.126 & 0.088 & 0.513 & 0.502 \\
    MantraNet (Wu et al. 2019) &0.051 & 0.018 & 0.500 & 0.500\\
    SPAN \cite{Span:eccv2020}  & 0.160 & 0.045 & 0.510 & 0.498 \\
    HiFi\_IFDL \cite{Hierarchical:cvpr2023} &0.145 &0.115 &0.502 &0.502 \\
    PSCC-Net \cite{pscc:csvt2022} & 0.208 & 0.118 & 0.514 & 0.505\\
    CAT-Net \cite{Catnet:wacv2021} & 0.301 & 0.194 & 0.589 & 0.537\\
    MVSS-Net \cite{mvss:iccv2021} & 0.234 & 0.153 & 0.568 & 0.515\\
    IF-OSN \cite{ison:tifs2022}  & 0.184 & 0.103 & 0.516 & 0.522\\
    Ours & \textbf{0.444} & \textbf{0.335} & \textbf{0.677} & \textbf{0.545}\\
    \bottomrule
  \end{tabular}
}
\caption{Evaluation results for image-editing based methods using CIMD-R. Pixel-level F1 scores are calculated using both best and fixed (0.5) thresholds. For image-level performance, AUC and image-level accuracy are reported.
}
\label{tab:CIMD-R}
\end{table}

For original Y-channel DCT coefficients $Q_{0}$, we perform a clipping operation using a threshold value T, after which we convert them into a binary volume. Denote this binary value conversion as $f: Q_{0}^{H\times W}\rightarrow  \left \{ 0, 1 \right \}^{(T+1) \times H \times W}$. It is shown in \cite{binary_volume:spl2020} that $f$ is effective in evaluating the correlation between each coefficient in the DCT histogram. Therefore, the DCT coefficients $Q_{0}$ is converted to binary volumes as:
\begin{equation}
f(Q_{0}^{t}(i, j)) = \left\{
             \begin{array}{ll}
             1 , & \text{if} \left |clip(Q_{0}(i, j))  \right | = t, t \in\left [ 0, T \right ],  \\
             0, & \text{otherwise}.
             \end{array}
\right.
\nonumber
\end{equation}
The function $clip(\cdot)$ is utilized to extract the histogram feature within $[-T, T]$, which is essential for GPU memory constraints. We set $T$ as 20 from the experiments. Additionally, we apply the absolute operation as DCT histogram exhibits symmetry.

\begin{table}[t]
\centerline{
\setlength{\tabcolsep}{1.2mm}
\footnotesize
  \begin{tabular}{lcccc}
    \toprule
    \multirow{2}{*}{Method} &
      \multicolumn{2}{c}{Pixel-level F1} &
      \multicolumn{2}{c}{Image-level} \\
      & Best & Fixed & AUC & Acc\\
      \midrule
    DJPEG \cite{double_compression_detection:2018} & 0.026 & 0.022 & 0.500 & 0.500  \\
    Comprint \cite{comprint:2022} & 0.030 & 0.010 & 0.467 & 0.500  \\
    CAT-Net \cite{Catnet:wacv2021} & 0.395 & 0.259 & 0.534 & 0.490  \\
    Ours & \textbf{0.542} & \textbf{0.442} & \textbf{0.727} & \textbf{0.525}  \\
    \bottomrule
  \end{tabular}
}
\caption{Evaluation results for compression-based methods on the CIMD-C subset. 
}
\label{tab:CIMD-C}
\end{table}

The compression artifact learning method involves two element-wise multiplication operations. The first multiplication is performed between the histogram features and the Q-matrix, which is utilized to simulate the JPEG de-quantization procedure. The second multiplication is used to guide the histogram feature to focus more on unstable coefficients, which is a critical step for detecting double-compressed images using the same QF.

In an $8 \times 8$ block of DCT coefficients, each coefficient position represents a specific frequency component. However, the convolution operations in the backbone are designed for RGB images and ignore these frequency relationships. To fully exploit the spatial and frequency information of the DCT coefficients, a reshaping operation is necessary. In detail, each block with a size of ($8 \times 8 \times 1$) is reshaped into a size of ($1 \times 1 \times 64$). Thus, the first and second dimensions represent the spatial information, while the third dimension represents the frequency relationship. Next, the de-quantized, quantized, and residual histogram features are concatenated in the channel dimension. Finally, the concatenated features are input to a $1 \times 1$ convolutional layer and the backbone network for the detection task.


\section{Experimental Results}

We first describe the experimental setup, and then compare the proposed network with the state-of-the-art methods on the newly proposed CIMD dataset.


\textbf{Datasets}. The training datasets used in this study were adopted from \cite{Catnet:wacv2021}. 
The testing phase entailed the utilization of CIMD-R and CIMD-C to evaluate the efficacy of image-editing-based and compression-based methods, respectively. 

\textbf{Evaluation metrics}. Following most previous work, we evaluated the localization results using pixel-level F1 score with both the optimal and fixed 0.5 thresholds. For image-level detection, we employed AUC and image-level accuracy. We set 0.5 as the threshold for image-level accuracy. Only tampered images are used for the manipulation localization evaluation.

\textbf{Implementation details}. Our model was implemented using PyTorch \cite{pytorch:nips2019} and trained on 8 RTX 2080 GPUs, with batch size 4. We set the initial learning rate as 0.001 with exponential decay.
The training process consists of 250 epochs. The proposed model is designed to accept various image formats, including both JPEG and non-JPEG formats. The training objective is designed to minimize the pixel-level binary cross-entropy. 

\subsection{Comparison With State-of-the-Art}
\label{sec:SoTA}

To guarantee a fair comparison and evaluate the previous models using newly introduced CIMD, we select the state-of-the-art approaches using these two standards: (1) pre-trained model is publicly available, and (2) the evaluation datasets we used are not in their training sets. Following these criteria, we select RRU-Net, MantraNet, HiFi\_IFDL, CR-CNN, SPAN, PSCC-Net, MVSS-Net, IF-OSN, CAT-Net, DJPEG and Comprint. 
All the work we compared are appropriately referenced in the related work section. 
We use CIMD-R to evaluate the performance of the image-editing-based method, while CIMD-C is utilized for compression-based approaches. 

\textbf{Evaluation using CIMD-R subset}. Table~\ref{tab:CIMD-R} reports the results of image-editing-based methods using CIMD-R, in which all image samples are uncompressed. Two Pixel-level F1 scores are calculated using the best F1 threshold for each image and using fixed F1 threshold of 0.5, respectively. Best scores are highlighted in bold. 
Our method outperforms existing SoTA methods in both image-level and pixel-level evaluation, which demonstrates its superiority for detecting small tampering regions.

\textbf{Evaluation using CIMD-C subset}. Table~\ref{tab:CIMD-C} compares the performance of compression-based IMD methods, where all image samples are double compressed using the same QF and the evaluation settings are consistent with those used in Table~\ref{tab:CIMD-R}. Our method is again the best performer in terms of overall performance, highlighting the effectiveness of our approach for double-compressed images with the same QF.

{\bf Ablation study}. We provide a simple ablation study shown in Table~\ref{tab:ablation}. Observe that our RGB stream is effective in both compressed and uncompressed data. Notably, the frequency stream fails to produce satisfactory results in CIMD-R due to the absence of compression artifacts. However, when the two branches work collaboratively, the model's performance improves in both localization and detection evaluation. 


\begin{table}[t]
\centerline{
  \footnotesize
  \begin{tabular}{lcccc}
    \toprule
    \multirow{2}{*}{Method} &
      \multicolumn{2}{c}{CIMD-R Subset} &
      \multicolumn{2}{c}{CIMD-C Subset} \\
      & F1 & AUC & F1 & AUC \\
    \midrule
    RGB Stream        & 0.330 & 0.593 & 0.409 & 0.525 \\
    Frequency Stream  & 0.130 & 0.531 & 0.301 & 0.512 \\
    RGB + Freqnency   & \textbf{0.335} & \textbf{0.677} & \textbf{0.442} & \textbf{0.727}  \\
    \bottomrule
  \end{tabular}
} 
\caption{Ablation study of two streams to work collaboratively and/or separately.}
\label{tab:ablation}
\end{table}

\section{Conclusion}

This study presents a novel Challenging Image Manipulation Detection (CIMD) dataset, which comprise of two subsets that are designed for evaluating image-editing-based and compression-based approaches, respectively. The datasets were manually taken and tampered with, and come with high-quality annotations. Additionally, we propose a two-branch method that outperforms state-of-the-art models in detecting image manipulations using the CIMD dataset. We have released our dataset to facilitate future research.



\section{Ethics Statement}
\label{sec:ethics}

To ensure ethical compliance, all photos presented in our dataset are original and obtained either in public places or with the owners' explicit permission in private places, in accordance with local jurisdiction laws. Moreover, the authors ensure that the photos contain neither identifiable individuals nor personal information. As advised by institutional review boards (IRB), IRB approval is not required for the dataset.

\bibliography{aaai24}

\end{document}